\documentclass[journal]{IEEEtran}

\usepackage{amsmath,amsfonts}
\usepackage{algpseudocode}
\usepackage{algorithmicx,algorithm}
\usepackage{hyperref}
\usepackage{array}
\usepackage[caption=false,font=normalsize,labelfont=sf,textfont=sf]{subfig}
\usepackage{textcomp}
\usepackage{stfloats}
\usepackage{url}
\usepackage{verbatim}
\usepackage{graphicx}
\usepackage{cite}

\usepackage[square,sort&compress,numbers]{natbib}
\usepackage{amsmath}
\usepackage{amsfonts} 
\usepackage{amssymb} 
\usepackage{multirow}
\usepackage{threeparttable}
\usepackage{booktabs}

\usepackage{bbm, dsfont}
\usepackage{bbding}
\usepackage{ifsym}

\graphicspath{{figs/}}
\graphicspath{{bio/}}

\usepackage{color, xcolor}

\renewcommand\thesection{\arabic{section}} 
\renewcommand\thesubsection{\arabic{section}.\arabic{subsection}} 
\renewcommand\thesubsubsection{\arabic{section}.\arabic{subsection}.\arabic{subsubsection}} 

\begin{document}
\title{GS2Pose: Two-stage 6D Object Pose Estimation Guided by Gaussian Splatting}

\author{
Jilan~Mei,~Junbo~Li,~Cai~Meng$^\star$
\\
\vspace{4pt}
Beihang University
}

\maketitle

\setlength{\abovedisplayskip}{6pt}
\setlength{\belowdisplayskip}{2pt}

\begin{abstract}
\newcommand{\blue}[1]{\textcolor{blue}{#1}}
This paper proposes a new method for accurate and robust 6D pose estimation of novel objects, named GS2Pose. By introducing 3D Gaussian splatting, GS2Pose can utilize the reconstruction results without requiring a high-quality CAD model, which means it only requires segmented RGBD images as input. Specifically, GS2Pose employs a two-stage structure consisting of coarse estimation followed by refined estimation. In the coarse stage, a lightweight U-Net network with a polarization attention mechanism, called Pose-Net, is designed. By using the 3DGS model for supervised training, Pose-Net can generate NOCS images to compute a coarse pose. In the refinement stage, GS2Pose formulates a pose regression algorithm following the idea of reprojection or Bundle Adjustment (BA), referred to as GS-Refiner.  By leveraging Lie algebra to extend 3DGS, GS-Refiner obtains a pose-differentiable rendering pipeline that refines the coarse pose by comparing the input images with the rendered images. GS-Refiner also selectively updates parameters in the 3DGS model to achieve environmental adaptation, thereby enhancing the algorithm's robustness and flexibility to illuminative variation, occlusion, and other challenging disruptive factors. GS2Pose was evaluated through experiments conducted on the LineMod dataset, where it was compared with similar algorithms, yielding highly competitive results. The code for GS2Pose will soon be released on GitHub.

\end{abstract}


\IEEEpeerreviewmaketitle
\section{Introduction}

%

\IEEEPARstart {A}ccurate 6D object pose estimation is a fundamental problem in the field of computer vision, with broad application prospects in technologies such as robot navigation\cite{deng2020self,cai2024gs} and virtual reality\cite{marchand2015pose,su2019deep}. However, classical pose estimation algorithms\cite{lowe1999object,lepetit2005monocular,hinterstoisser2013model} lack robustness against environmental interference, such as non-uniform lighting, varying degrees of occlusion, and dynamic blur. Moreover, the lightweight nature of the algorithm is also demanding in the field of embodied intelligence\cite{huang2023voxposer,driess2023palm,long2024discuss}.

With the widespread application of deep learning methods, the robustness of related algorithms\cite{lin2024sam,li2018deepim,li2019cdpn,he2020pvn3d,he2021ffb6d} against interference has continually improved. Early works\cite{su2022zebrapose,wang2019densefusion,wang2021gdr,xiang2017posecnn} have achieved high-precision instance-level pose estimation. However, these models can only handle a specific object after the training session and cannot generalize to others. Additionally, they require datasets with precise ground truth poses, which are difficult to obtain in practical applications.

The emergence of novel pose representation methods \cite{wang2019normalized}, such as NOCS, has led to breakthroughs in category-level pose estimation methods\cite{song2016deep,qi2018frustum,chen2016monocular,mousavian20173d,xiang2015data,tian2020shape}, achieving notable intra-class generalization. Trained models can perform high-precision pose estimation on objects with similar geometric and color features. However, these methods typically require a substantial number of  CAD models of the same category during the training phase, results in huge time expenditure. Additionally, since the 6D pose of the target object is bound to the objects coordinate system under the CAD model, which can lead to issues, such as parameter ambiguity in the estimation results during the inference phase.

In recent years, with the development of large models\cite{leiter2024chatgpt,kirillov2023segment,zhao2023survey}, some research\cite{nguyen2024gigapose,labbe2022megapose,wen2024foundationpose} have introduced the concept of pre-training on large datasets into the field of 6D pose estimation. These methods construct large datasets by collecting numerous CAD models of common objects from different categories, enabling effective generalization to unseen objects. They require only the CAD model of the target object during inference, allowing for the artificial setting of strict coordinate relationships without the need for additional training on the target object. However, these models also have drawbacks, such as the inability to generalize to uncommon objects, high consumption of computational resources, and their accuracy being heavily dependent on the quality of CAD modeling.

To address the aforementioned shortcomings of existing algorithms, a novel pose estimation method is proposed that eliminates the need for artificially designed CAD models. This method is designed for application scenarios where high-quality CAD models of the target object are unavailable, and only untextured scanned models or structure-from-motion (SFM) point cloud models can be obtained. To achieve lightweight training, accurate reference relationships, and robustness to interference, GS2Pose consists of a two-stage pose estimation approach comprising coarse estimation followed by pose refinement.

\begin{figure*}[!th]
\centering
\resizebox{0.99\linewidth}{!}{
\includegraphics[width=\linewidth]{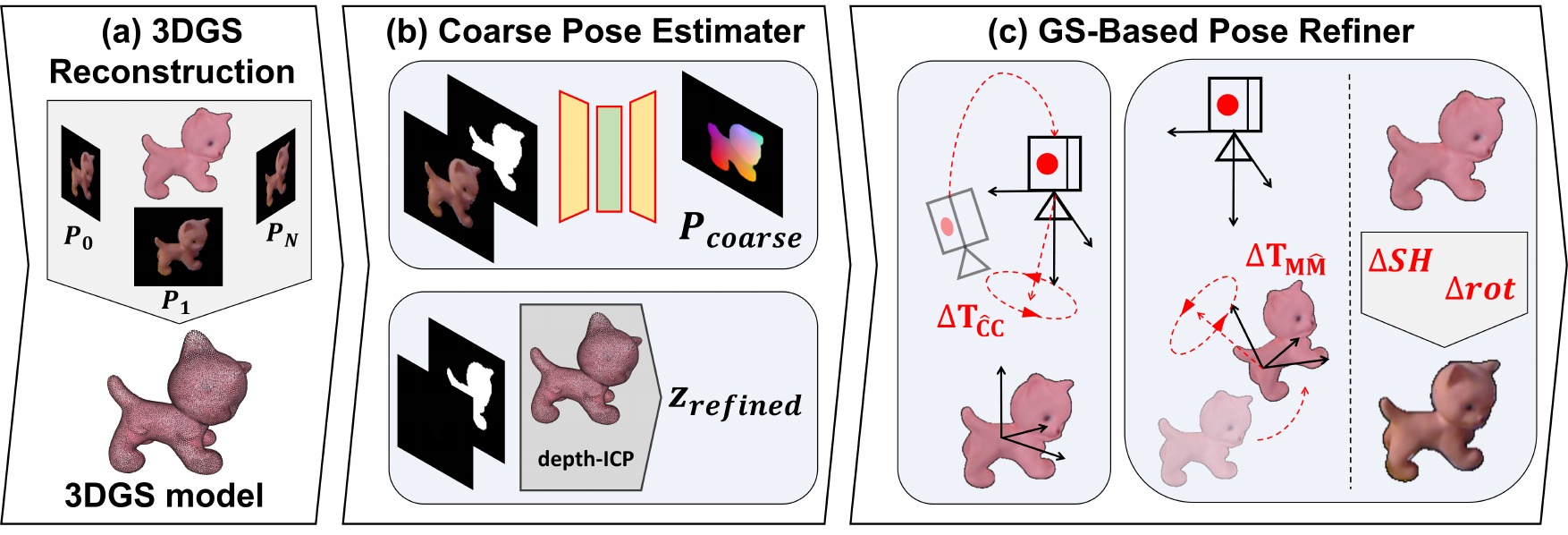}
}
\caption{The structure of the GS2POSE
}
\label{fig:model}
\end{figure*}

The detailed process of GS2Pose is illustrated in Fig. \ref{fig:model}. The 3DGS point cloud model of the object (hereafter referred to as the 3DGS model) is obtained using existing 3DGS reconstruction techniques, with the object coordinate system manually specified. Utilizing insights from the GS-SLAM model\cite{yan2024gs}, the commonly used reprojection-based pose optimization iterative approach from the SLAM domain is introduced \cite{engel2014lsd,mur2017orb,keetha2024splatam}, also known as Bundle Adjustment (BA). By representing object poses using Lie algebra and integrating this representation with the differentiable 3DGS rendering pipeline, an approach is implemented that utilizes reprojection and backpropagation. This enables an iterative optimization algorithm that can regress both the object pose and the camera pose, referred to as GS-Refiner. 

Since the iterative optimization algorithm requires a reasonable initial pose as a starting point, it is necessary to design an algorithm that can provide a rough pose estimate based solely on the segmented object image. Inspired by the NeRF-Pose model\cite{li2023nerf}, a rough pose estimation network named Pose-Unet was developed. RGB images and their corresponding NOCS images are obtained from the camera perspective using 3DGS. These images are subsequently input into a pre-trained coarse pose estimation network (Pose-Unet) for fine-tuning, resulting in a coarse pose estimation for any novel rendering view of the object. 

On the other hand, GS-Refiner leverages the parameter interpretability of the 3DGS model to selectively optimize and refine parameters, such as higher-order spherical harmonic color parameters, transparency, and ellipsoid orientation through backpropagation. This allows the surface colors to adaptively adjust to environmental factors encountered during actual capture, such as lighting, occlusion, and motion blur.

\vspace{6pt}
The primary contributions of the paper can be summarized as follows:

i) By incorporating 3DGS reconstruction technology, lightweight 6D pose estimation of previously unseen objects is achieved in the absence of CAD models.

ii) By employing Lie algebra to modify the differentiable rendering pipeline of 3DGS, a reprojection iterative algorithm called GS-Refiner has been developed developed, enabling the correction of both object poses and camera poses.

iii) By selectively regressing the parameters of 3DGS, a 6D pose estimation algorithm was developed with robust resistance to complex lighting, motion blur, and occlusions.

iv) Through experiments on datasets such as LineMod, the GS2Pose model demonstrated substantial advantages over comparable algorithms, particularly in terms of accuracy, inference speed, and computational resource efficiency.

\vspace{6pt}
\section{Related Works}
\newcommand{\blue}[1]{\textcolor{blue}{#1}}
\renewcommand{\thesection}{\arabic{section}}  
\renewcommand{\thesubsection}{\thesection.\arabic{subsection}}

%
%

This section provides a brief summary of the development on the 6D pose estimation. We first review the 6D pose prediction about known rigid objects. Then we focus on the progress of 6D pose estimation about novel objects in recent years. We summarize the recent development of Gaussian models finally.

\subsection{6D pose estimation of seen objects}

Traditional 6D pose estimation methods\cite{lowe1999object,bay2006surf,collet2010efficient,hinterstoisser2011multimodal} rely on extracting local invariant features and establishing correspondences by template matching. Researchers have made innovative explorations in the features robustness and the template matching performance in complex occlusion scenarios. However, these traditional methods still struggle to solve challenges related to large variations in lighting and the accurate pose estimation of symmetric objects. As a result, the 6D pose estimation becomes inefficient and unsuitable for widespread development and practical applications.

Conversely, deep learning methods have gained attention in 6D pose estimation due to their powerful ability to automatically learn features from datasets. The PoseCNN model\cite{xiang2017posecnn} introduced a novel loss function, enabling the network to better handle symmetric objects, thereby enhancing the robot's ability to interact with the real world. As for the applications without depth information, BB8\cite{rad2017bb8} model proposed a classifier to restrict the range of poses, which can compensates the lack of depth information. Moreover, RADet\cite{hai2023rigidity} proposed a rigidity-aware detection method to better address occlusion issues, which created a visibility map using the minimum barrier between each pixel in the detection bounding box and the box boundary.

Recently, Generative Adversarial Networks (GAN) have demonstrated exceptional capabilities in denoising and recovering missing parts of images. UnrealDA\cite{zakharov2018keep} proposed a GAN-based network, which transformed real depth maps with background occlusion into synthetic depth maps to improve pose estimation performance. Apart from that, the Pix2Pose\cite{park2019pix2pose} model based on GAN network, introduced a transformer loss to guide predictions toward the closest pose, addressing pose estimation for symmetric objects.

\subsection{6D pose estimation of unseen objects}

To improve the generalization ability and robustness of 6D pose estimation with CAD models, some researchers aim to address pose estimation for novel objects. MegaPose\cite{labbe2022megapose} network proposed a 6D pose estimator based on a rendering and comparison strategy, which trains the network on a large synthetic dataset. Moreover, GigaPose\cite{nguyen2024gigapose} network proposed a novel solution by leveraging templates to recover out-of-plane rotations, then utilizing patches correspondences to estimate the four remaining pose parameters. Although above foundation methods have strong generalization capabilities, their robustness remains insufficient for specialized devices in industries and medical, such as surgical instruments and precision constructions. We proposed a course-refine 6D pose estimation network. For each new object, coarse estimation network provides an approximate pose by rapid training, followed by precise correction in the refine estimation network.


\subsection{6D pose estimation with 3D reconstruction model}

3DGS\cite{kerbl20233d} demonstrates significant advantages in high-quality and real-time rendering. This work represents scene with 3D Gaussian ellipsoids and efficiently renders by rasterizing the Gaussian ellipsoids into images, achieving state-of-the-art (SOTA) level visual quality. At the same time, 3DGS employs an explicit construction method, possessing clear geometric structure and appearance. This technology has already been applied in multiple fields, including autonomous navigation\cite{yan2024gs,keetha2024splatam,matsuki2024gaussian,yugay2023gaussian}, virtual human body reconstruction\cite{chen2024monogaussianavatar,wang2023gaussianhead} and 3D generation\cite{chen2024text,tang2023dreamgaussian,yi2023gaussiandreamer}.


However, there are few works that apply 3D Gaussian Splatting (3DGS) to the field pose estimation currently. Although GSPose network attempts to apply 3DGS model, it still requires training a DINO network to create a database, which rely on dataset training. So this research is difficult to fine-tune for new objects. Moreover, it does not fully utilize the differentiable advantages of 3D Gaussian.

\section{Methodology}

\subsection{Overview}

In this chapter, we provide a detailed overview of the framework and principles of pose estimation methods. Our objective is to determine the relative pose of an object with respect to the camera, based on the input RGB-D image $I_{\text{in}}$ and the 3D geometric reconstruction model $G_m$ of the object. This involves computing the transformation matrix \(T_{cm}\) from the coordinate system of the reconstructed model \(m\) to the camera coordinate system \(c\), which is composed of a translation vector \(t_{cm}\) and a rotation matrix \(R_{cm}\).

\begin{align}
\begin{split}
T_{cm} &= \left[ \begin{array}{cc}
R_{cm} & t_{cm} \\
0^{T} & 1
\end{array} \right], \quad 
t_{cm} = \begin{pmatrix} x_{cm} & y_{cm} & z_{cm} \end{pmatrix}^{T}
\end{split}
\end{align}

To achieve the aforementioned objective, we first reconstruct the 3D Gaussian Splatting (3DGS) model of the target object. Subsequently, under the supervision of this 3DGS model, we train a coarse estimation network, referred to as Pose-net, which is capable of generating NOCS images from novel viewpoints and predicting the coarse pose \(T_{cm}^{\mathrm{coarse}}\) of the object in RGB images captured from arbitrary angles. Finally, we propose a novel refinement algorithm that utilizes the coarse predicted pose as an initial estimate, following an iterative optimization approach based on 3DGS reprojection. By continuously minimizing the differences between the rendered images and the input images, we refine and optimize the pose to obtain an accurate final output \(T_{cm}^{\mathrm{refined}}\).


\subsection{3D Gaussian Splatting}

3D Gaussian Spheres (3DGS) is a scene representation method that describes objects in the world coordinate system using Gaussian spheres. All attributes of the 3D Gaussian Spheres are learnable, including the position parameters \( \mu \), opacity \( a \), the 3D covariance matrix \( r \), and the spherical harmonics \( sh \). Given any point \( \mathbf{x} \) in the world coordinate system, the 3D Gaussian sphere defined at point \( \mathbf{x} \) according to the Gaussian distribution is as follows:
\begin{align} 
\begin{split}
f(\mathbf{x}; \mu, \Sigma) = \exp \left( -\frac{1}{2} (\mathbf{x} - \mu)^{\mathrm{T}} \Sigma^{-1} (\mathbf{x} - \mu) \right)
\end{split} 
\end{align}

\begin{align} 
\begin{split}
\Sigma = R S S^{\mathrm{T}} R^{\mathrm{T}}
\end{split} 
\end{align}

where \( R \) denotes the rotation matrix computed from \( r \), and \( S \) represents the diagonal matrix derived from \( s \). Subsequently, a fast rasterization approach is employed to project the 3D Gaussian points onto a 2D plane for rendering.


\subsection{Coarse Pose Estimation Network}

Inspired by the NeRF-Pose model\cite{li2023nerf}, which is currently the state-of-the-art approach in 6D pose estimation, we have designed a lightweight NOCS image generation network ( Pose-Unet ) to predict the coarse pose of objects.  The 3DGS method generates RGB images from the camera viewpoint along with the corresponding NOCS images, which are used as training inputs for Pose-Unet. Through fine-tuning, the model can rapidly generalize to new objects. Subsequently, the test RGB images (segmented using the CNOS model) are input to obtain the corresponding NOCS images, from which a coarse pose is estimated. Since the NOCS image predictions exhibit significant deviations along the z-axis, the improved ICP algorithm is utilized to align the point cloud model in the observed viewpoint (acquired from RGB-D images) with the Gaussian model, correcting the z-axis in the coarse pose.

Pose-Unet utilizes ResNet50 as the encoder. While in the decoder stage, three transposed convolution layers are employed for up sampling. As most encoder-decoder based network models, Pose-Unet incorporates skip connections ( Mobile-ASPP)  during the down sampling to minimize information loss. Mobile-ASPP optimizes the ASPP structure, which consists of three parallel atrous convolutions. Specifically, the dilated convolution layers have kernel sizes of $1 \times1$, $3 \times3$, and $3 \times3$, with corresponding dilation rates of 1, 1, and 2, respectively.  This module enables the network to fully capture shallow information while reducing computational resource consumption.  In the deep feature extraction module, based on the PPM structure, four parallel pooling layers with different kernel sizes are constructed to effectively capture dependencies between pixels.


A deep estimation method for point cloud registration on object surfaces is proposed. The point cloud $P_1$ is generated by combining the RGB image and depth image from the target viewpoint. By combining the CAD model with camera pose estimation, the point cloud $P_2$ of the object’s surface facing the camera is generated. By calculating the average z-values, $Z_1$ and $Z_2$, of the two point clouds $P_1$ and $P_2$ in the camera coordinate system along the z-axis, the estimated $Z_2$ is corrected to $Z_1$. This completes a pose correction along the depth direction.

\begin{figure}[!t]
\centering
\resizebox{0.99\linewidth}{!}{
\includegraphics[width=\linewidth]{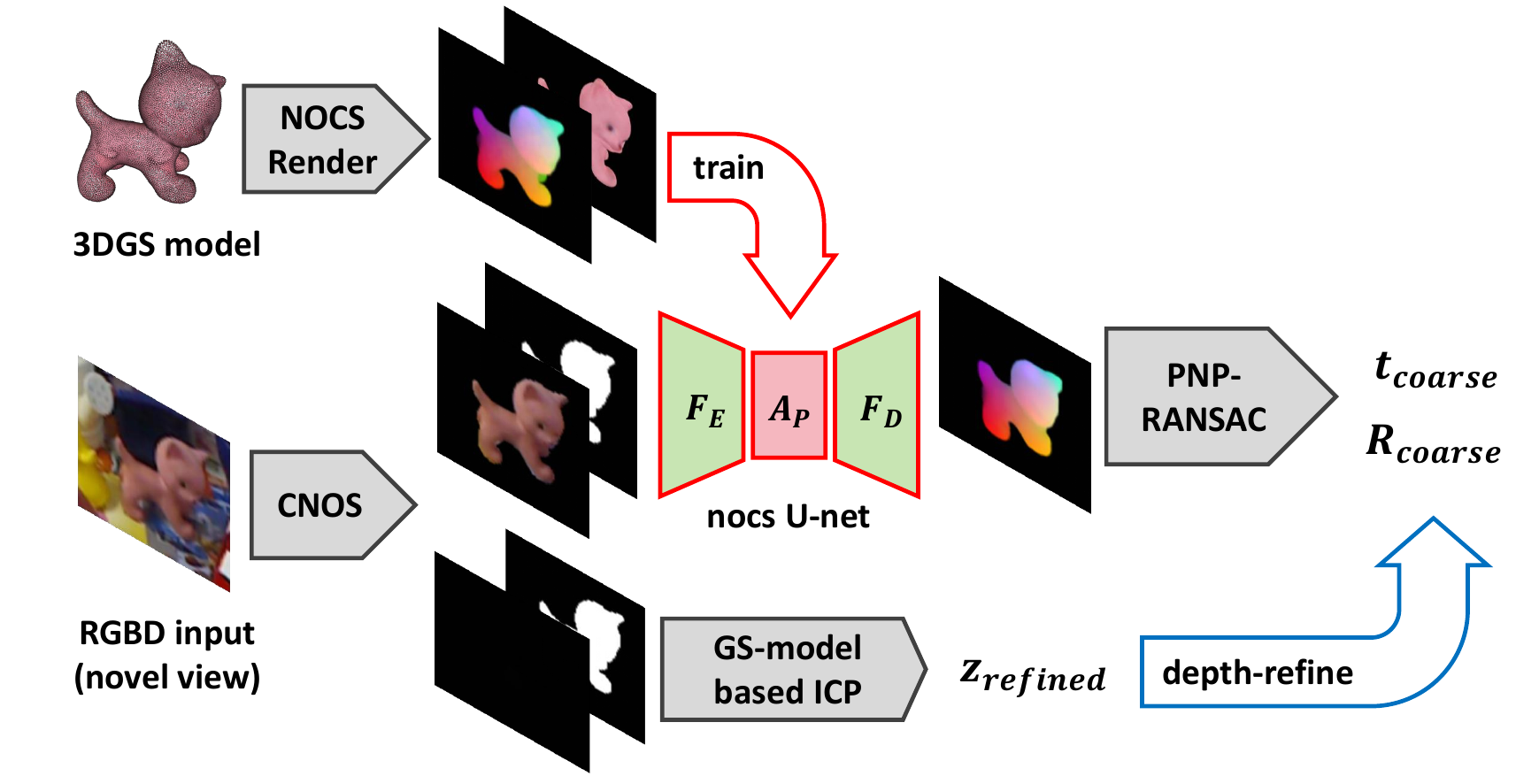}
}
\caption{The structure of the Pose-Unet
}
\label{fig:coarse network}
\vspace{-8pt}
\end{figure}

\subsection{Refine Pose Estimation Network}
\subsubsection{Overview}
After obtaining a coarse estimation \(T_{cm}^{\mathrm{coarse}}\) with limited accuracy, we designed a multi-stage refinement algorithm, termed GS-refiner, which leverages the 3DGS representation model of the object. This algorithm employs an iterative reprojection method to provide a precise pose estimation of the object.

Inspired by 3D Gaussian Splatting SLAM\cite{yan2024gs}, we represent the pose changes between coordinate systems using Lie algebra. We compute the error through reprojection for backpropagation, aiming to regress the precise pose of the object.

Thanks to the differentiable rendering pipeline of 3DGS, we can differentiate most parameters of the 3DGS, including the rendering pose, by calculating the differences between the reprojection images \(I_{\mathrm{pred}}(T_{cm}^{\mathrm{iter}})\) under the coarse estimated pose and the ground truth images \(I_{\mathrm{in}}\). Following the approach of 3D Gaussian Splatting, we design the loss function as follows:
\begin{align}
\begin{split}
\text{loss}(I_{\mathrm{in}}, I_{\mathrm{pred}}) &= \lambda L_{1} + (1 - \lambda) L_{\mathrm{DSSIM}}
\end{split}
\end{align}

where \(\lambda\) is a hyperparameter, \(\mathcal{L}_{1}\) represents the L1 loss between two images, and \(\mathcal{L}_{D-\mathrm{SSIM}}\) represents the D-SSIM loss. Specifically, let the loss at a certain pixel \(p(u,v)\) be determined by the value of that pixel:
\begin{align}
\begin{split}
L(u,v) &= \text{loss}(p_{\mathrm{pred}}, p_{\mathrm{gt}})
\end{split}
\end{align}

which is influenced by the 2D elliptical projections of multiple 3D Gaussian Splatting ellipsoids projected onto it. This can be expressed using the ray casting formula:
\begin{align}
p_{\mathrm{pred}} &= \sum_{i=1}^{N} c_{i} \alpha_{i} \prod_{j=i-1}^{1} (1 - \alpha_{j}) 
\end{align}

The RGB color vector \( c_{i} \) can be obtained from the spherical harmonic parameters of the 3DGS ellipsoid and the relative pose \( T_{cm} \) with respect to the camera. The transparency \( \alpha_i \) is determined by the distance between the current pixel and the center point \( p_{i} \) of the 2D ellipse projection, as well as the Gaussian covariance parameters \( \Sigma_{i} \) of the projected ellipse. Furthermore, the center point \( p_{i} \) is determined by the camera's relative pose \( T_{cm} \), namely:
\begin{align}
p_{i} = \pi(T_{cm}, p_{m})
\end{align}

Based on the \( T_{cm} \) and the camera intrinsic parameter matrix \( K \), a Jacobian matrix \( J \) can be generated for the purpose of flattening the ellipsoid parameters into the plane. \( \Sigma_{i} \) can be determined using the Jacobian matrix \( J \) and the rotational part of the relative pose \( R_{cm} \):
\begin{align}
\Sigma_{i} &= J \Sigma_{c} J^{\mathrm{T}} \\
\Sigma_{c} &= R_{cm} \Sigma_{m} R_{cm}^{\mathrm{T}}
\end{align}

according to the chain rule of differentiation:
\begin{align}
\frac{\partial p_{i}}{\partial T_{cm}} &= \frac{\partial p_{i}}{\partial p_{c}} \frac{\partial p_{c}}{\partial T_{cm}} \\
\frac{\partial \Sigma_{i}}{\partial T_{cm}} &= \frac{\partial \Sigma_{i}}{\partial J} \frac{\partial J}{\partial p_{c}} \frac{\partial p_{c}}{\partial T_{cm}} + \frac{\partial \Sigma_{i}}{\partial R_{cm}} \frac{\partial R_{cm}}{\partial T_{cm}} \\
\frac{\partial c_{i}}{\partial T_{cm}} &= \frac{\partial c_{i}}{\partial t_{cm}} \frac{\partial t_{cm}}{\partial T_{cm}}
\end{align}

Due to the discontinuity of the matrix form of \( T_{cm} \) in \( \mathbb{R}^{4 \times 4} \), \( \frac{\partial p_{c}}{\partial T_{cm}} \) and \( \frac{\partial R_{cm}}{\partial T_{cm}} \) cannot be directly differentiated. Therefore, we need to convert \( T_{cm} \) into the Lie algebra form before performing the differentiation.

Let the homogeneous coordinates of any point in the point cloud model in the object coordinate system be denoted as \( p_{m} = [x_{m}, y_{m}, z_{m}, 1]^{\mathrm{T}} \) and the homogeneous coordinates in the camera coordinate system be denoted as \( p_{c} = [x_{c}, y_{c}, z_{c}, 1]^{\mathrm{T}} \). When the non-homogeneous form of the coordinates is used, it will be indicated by a subscript, such as \( p_{c}^{:3} \). According to the definition of the transformation matrix, we have:
\begin{align} 
\begin{split}
p_{c} = T_{cm} p_{m}
\end{split} 
\end{align}

It is important to note that, based on the knowledge of Lie algebra, altering the camera pose in the object coordinate system (perturbing the shooting perspective) yields fundamentally different effects compared to changing the object pose in the camera coordinate system when refining relative poses \(T_{cm}\). Subsequent formula derivations and experiments demonstrate that these two approaches efficiently correct the translational and angular relationships of the object relative to the camera.

Consequently, we have separated the Refiner into two components: perspective pose correction (Camera refiner) and object pose correction (Object refiner). Below, we will introduce the principles of these two components, provide a brief derivation of the relevant formulas, and finally explain how we integrate these two components to form the GS Refiner.

\subsubsection{Camera Refiner} In the Camera Refiner, the object being updated is the camera coordinate system \(c\). During each iteration, a new camera coordinate system \(c'\) is obtained, and the coordinates of any object point \(p_m\) in the new camera coordinate system are given by:

\begin{align}
p_{c'} = T_{c'c} T_{cm} p_m = T_{c'c} p_c
\end{align}

Here, \( T_{c'c} \in SE(3) \) can be viewed as a left perturbation applied to \( T_{cm} \). Let the Lie algebra corresponding to \( T_{c'c} \) be denoted as:
\begin{align} 
\begin{split}
\tau_c = \begin{bmatrix} \rho_c & \varphi_c \end{bmatrix}^T \in \mathfrak{se}(3) \quad p_{c'} = \exp(\tau_c) p_c
\end{split} 
\end{align}

Since the rendered coordinate system at this point is the transformed camera coordinate system, that is:
\begin{align} 
\begin{split}
\frac{\partial p_i}{\partial T_{cm}} = \frac{\partial p_i}{\partial p_c} \cdot \frac{\partial p_c}{\partial T_{cm}}
\end{split} 
\end{align}

Let \( p_{c'} \) take the derivative of \( \tau_c \), that is:
\begin{align}
\frac{\partial p_{c'}}{\partial \tau_c} = \left[ I, - p_{c'}^{3} \right]
\end{align}

On the other hand, the updated rotation matrix part is:
\begin{align} 
\begin{split}
R_{c'm} = R_{c'c} R_{cm}
\end{split} 
\end{align}

Where \( R_{c'c} \) corresponds to the Lie algebra \( \phi_c \), so we can obtain the derivative of the matrix \( R_{cm} \) with respect to \( \phi_c \):

\begin{align} 
\begin{split}
\frac{\partial R_{c'm}}{\partial \phi_c} = \left[ -R_{cm}^{:1}, -R_{cm}^{:2}, -R_{cm}^{:3} \right]^{T}
\end{split} 
\end{align}

Finally, from \( t_{c'} = (\phi_{c} + I) R_{cm} + \rho_{c} \), we can conclude that:
\begin{align} 
\begin{split}
\frac{\partial t_{c'm}}{\partial \rho_c} = I
\end{split} 
\end{align}

Through the above derivation, we have obtained \( \frac{\partial p_c}{\partial T_{cm}}, \quad \frac{\partial R_{cm}}{\partial T_{cm}}, \quad \frac{\partial t_{cm}}{\partial T_{cm}} \), which allows us to derive \( \frac{\partial p_i}{\partial T_{cm}}, \quad \frac{\partial \Sigma_i}{\partial T_{cm}}, \quad \frac{\partial c_i}{\partial T_{cm}} \), completing the construction of the back propagation chain and enabling the gradient descent update for the pose \( T_{cm} \).

\subsubsection{Object Refiner} In the second stage, the object of the update changes from the pose of the camera relative to the object to the pose of the object relative to the camera. Let the updated object coordinate system be \( m' \). Since each object point \( p_m \) on the object is rigidly attached to the object coordinate system, its coordinate values in the object coordinate system will not change, that is:
\begin{align} 
\begin{split}
p_m = p_{m'} \quad p_c = T_{cm} T_{mm'} p_m
\end{split} 
\end{align}

\( T_{mm'} \in SE(3) \) can be viewed as a right perturbation applied to \( T_{cm} \). Let the Lie algebra corresponding to \( T_{mm'} \) be denoted as:
\begin{align} 
\begin{split}
\tau_m = \begin{bmatrix} \rho_m & \varphi_m \end{bmatrix}^T \in se(3), \quad p_c = T_{cm} \exp(\tau_m) p_m
\end{split} 
\end{align}

By drawing an analogy to the derivation in Camera Refiner, we can obtain:
\begin{align}
\frac{\partial p_c}{\partial \tau_m} = \left[ T_{cm} \bullet I, - T_{cm} \bullet p_m^{:3} \right], \quad
\end{align}

\begin{align}
\frac{\partial R_{cm}}{\partial \phi_m} = \left[ - R_{cm_{1,:}}, - R_{cm_{2,:}}, - R_{cm_{3,:}} \right]
\end{align}

\begin{align}
\frac{\partial t_{c'm}}{\partial \rho_c} = R_{cm}
\end{align}

Through the derivation above, the backpropagation for Object Refiner has also been implemented. In the experiments, we will implement the aforementioned backpropagation chain using CUDA programming, enabling efficient computation for pose correction through the application of gradient descent algorithms.

\subsubsection{Environment adoption} Since the 3DGS model is a type of self-emissive model, the lighting and shading characteristics of the model are not derived from its relative pose to the light source. Instead, they are obtained through the superposition of the RGB colors of each Gaussian sphere. This is a distinctive feature of the RayCast rendering algorithm. To address issues such as reflections and shadows under varying lighting conditions, we leverage the learnable nature of the 3DGS color parameters and the anisotropic properties of color parameters expressed by spherical harmonics. This allows the model to adapt to changes in lighting while adjusting its pose, thereby enhancing the accuracy of the model and preventing angle miscorrections due to lighting or shadow issues.

In this step, we set the 16 spherical harmonic parameters of the Gaussian model as learnable parameters, along with the rotational pose parameter, rot, of the Gaussian spheres. We have observed that during the model's color learning process, there is a tendency for the back side of the Gaussian spheres to be assigned a black color. Allowing the Gaussian spheres to rotate freely can accelerate the learning efficiency and prevent overfitting of the colors, as well as mitigate the issue of vanishing gradients.

Additionally, we lock other parameters, such as the scale parameter of the Gaussian spheres, their position parameters (xyz) relative to the object coordinate system, and their transparency. This is to prevent the model from compromising its original structure during iterations in an attempt to forcefully fit the target image. Such compromises could negatively impact the accuracy of angle estimation.

By carefully managing these parameters, we ensure that the model retains its integrity while effectively adapting to various lighting conditions. This approach not only enhances the model's performance but also maintains the precision required for accurate angle calculations, ultimately leading to improved results in rendering.

\begin{figure*}[!th]
\centering
\resizebox{0.99\linewidth}{!}{
\includegraphics[width=\linewidth]{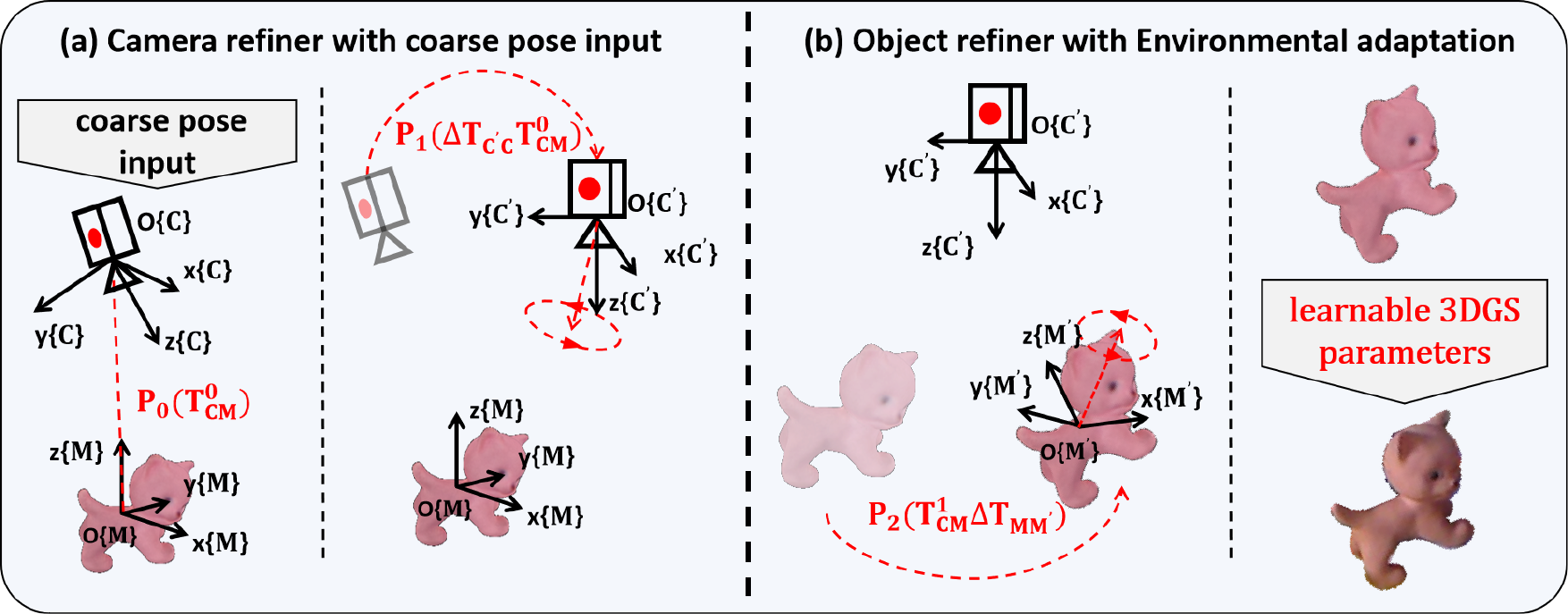}
}
\caption{The structure of the GS-Refiner
}
\label{fig:refine network}
\end{figure*}

\section{Experimental Results and Analyses}

In order to evaluate the effectiveness of the proposed model, this section conducts a comparative analysis of its performance against a range of state-of-the-art deep-learning 6D pose estimation models, including Pix2Pose \cite{park2019pix2pose}, SSD-6D \cite{kehl2017ssd}, Lienet \cite{karnati2021lienet}, Cai \cite{cai2020reconstruct}, DPOD \cite{zakharov2019dpod}, PVNet \cite{peng2019pvnet}, CDPN \cite{li2019cdpn}

\subsection{Experimental Dataset and Settings}
\label{sec:dataset}

Experiments were conducted on two publicly accessible datasets for 6D pose estimation: Linemod (LM) \cite{hinterstoisser2013model} and 

\vspace{3pt}
\noindent \textbf{Linemod (LM)} \cite{hinterstoisser2013model}: 
The LM dataset consists of 15 registered video sequences, each containing over 1100 frames. The object scales range from 100 mm to 300 mm.  There are significant variations in illumination intensity of the images captured under the same model, along with minimal occlusion phenomena. We referenced the majority of 6D pose estimation methods \cite{li2023nerf,labbe2022megapose} and selected 13 categories to evaluate the performance of the model, including ape, bvise, cam, can, cat, driller, duck, eggbox, glue, holep, iron, lamp and phone.



\begin{table*}[!thbp] 
\centering
\caption{
Comparison with other methods on the LineMOD test set. (ADD-0.1d)
}
\label{tab:LineMOD}
\resizebox{0.95\linewidth}{!}{
\begin{tabular}{c|c| *{13}{c}|c}
\toprule
Method 
& Publish
& $\text{ape}$ 
& $\text{bvise}$ 
& $\text{cam}$ 
& $\text{can}$ 
& $\text{cat}$ 
& $\text{driller}$ 
& $\text{duck}$ 
& $\text{eggbox*}$ 
& $\text{glue*}$ 
& $\text{holep}$
& $\text{iron}$
& $\text{lamp}$
& $\text{phone}$
& $\text{avg}$
\\
\midrule
SSD-6D\cite{kehl2017ssd} &[ICCV 2017] &  65.0 & 80.0 & 78.0 & 86.0 & 70.0 & 73.0 & 66.0 & \underline{100.0} & \textbf{100.0} & 49.0 & 78.0 & 73.0 & 79.0 & 76.7
\\
Pix2Pose\cite{park2019pix2pose} &[ICCV 2019] &  58.1 & 91.0 & 60.9 &	84.4 & 65.0 & 76.3 & 43.8 &	96.8 & 79.4 & 74.8 & 83.4 & 82.0 & 45.0 & 72.4
\\
DPOD\cite{zakharov2019dpod} &[ICCV 2019]  & 53.3 & 95.2 & 90.0 & 94.1 & 60.4 & \underline{97.4} & 66.0 & 99.6 & 93.8 & 64.9 & \underline{99.8} & 88.1 & 71.4 & 82.6
\\
CDPN\cite{li2019cdpn} &[ICCV 2019]  & 64.4 & 97.8 & 91.7 & 95.9 & 83.8 & 96.2 & 66.8 & 99.7 & \underline{99.6} & 85.8 & 97.9 & 97.9 & 90.8 & 89.9
\\
Cai\cite{cai2020reconstruct} &[CVPR 2020] &  52.9 & 96.5 & 87.8 & 86.8 & 67.3 & 88.7 & 54.7 & 94.7 & 91.9 & 75.4 & 94.5 & 96.6 & 89.2 &
82.9
\\
Lienet\cite{karnati2021lienet} &[TCDS 2022]  & 38.8 & 71.2 & 52.5 & 86.1 & 66.2 & 82.3 & 32.5 & 79.4 & 63.7 & 56.4 & 65.1 & 89.4 & 65.0 & 65.2
\\
PVNet\cite{peng2019pvnet} &[CVPR 2022]  & 43.6 & \textbf{99.9} & 86.9 & 95.5 & 79.3 & 96.4 & 52.6 & 99.2 & 95.7 & 81.9 & 98.9 & \underline{99.3} & 92.4 & 86.3
\\
OnePose\cite{sun2022onepose} &[CVPR 2022] &  11.8 & 92.6 & 88.1 & 77.2 & 47.9 & 74.5 & 34.2 & 71.3 & 37.5 & 54.9 & 89.2 & 87.6 & 60.6 & 63.6
\\
OnePose++\cite{he2022onepose++} &[NeurIPS 2022]  & 31.2 & 97.3 & 88.0 & 89.8 & 70.4 & 92.5 & 42.3 & 99.7 & 48.0 & 69.7 & 97.4 & 97.8 & 76.0 & 76.9
\\
NeRF-Pose\cite{li2023nerf} &[ICCVW 2023] &  69.4 & 99.4 & \textbf{98.3} & \textbf{97.8} & 77.8 & \textbf{99.6} & 69.7 & 99.9 & 98.9 & \underline{89.4} & \textbf{99.8} & \textbf{99.8} & \underline{94.8} & 91.8 
\\
GS-Pose\cite{cai2024gs} &[Arxiv 2024] &  \underline{71.0} & \underline{99.8} & \underline{98.2} & \underline{97.7} & \underline{86.7} & 96.2 & \underline{77.2} & 99.6 & 98.4 & 87.4 & 99.2 & 98.9 & 85.0 & \underline{92.0}
\\

\midrule
GS2pose &   & \textbf{95.4(24.4$\uparrow$)} & 99.2 & 97.4 & 95.1 & \textbf{93.2(6.5$\uparrow$)} & 95.3 & \textbf{83.8(6.6$\uparrow$)} & \textbf{100.0} & 98.4 & \textbf{93.5(4.1$\uparrow$)} & 94.7 & 93.9 & \textbf{95.0(0.2$\uparrow$)} & \textbf{95.0(3.0$\uparrow$)}
\\
\bottomrule
\end{tabular}
}
\end{table*}

\begin{figure*}[!th]
\centering
\resizebox{0.99\linewidth}{!}{
\includegraphics[width=\linewidth]{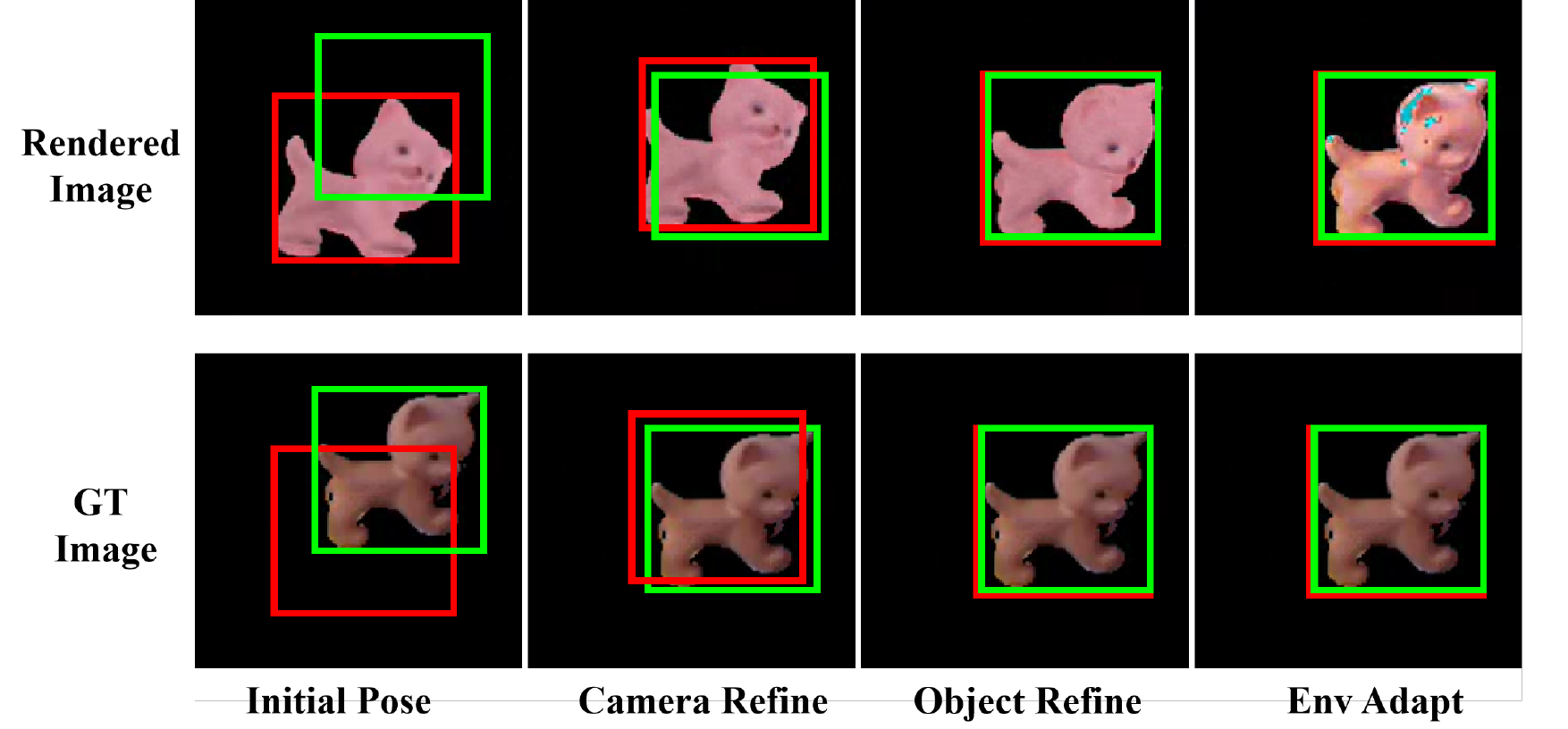}
}
\label{fig:refine network}
\end{figure*}


\section{Conclusion}

In conclusion, this paper presents GS2Pose, a novel method for accurate and robust 6D pose estimation of novel objects that effectively addresses the limitations of traditional approaches reliant on high-quality CAD models. By leveraging 3D Gaussian splatting and segmented RGBD images, GS2Pose demonstrates a significant advancement in the efficiency and accessibility of pose estimation. The two-stage architecture, comprising the coarse estimation via the Pose-Net and the refined estimation through the GS-Refiner, showcases a well-integrated approach that enhances the precision of pose estimation. The experimental results on the LineMod dataset confirm the effectiveness of GS2Pose, positioning it as a competitive alternative to existing algorithms in the field.

\ifCLASSOPTIONcaptionsoff
  \newpage
\fi

\bibliographystyle{IEEEtran}
\bibliography{IEEEabrv,myreferences}

\begin{thebibliography}{10}
\providecommand{\url}[1]{#1}
\csname url@samestyle\endcsname
\providecommand{\newblock}{\relax}
\providecommand{\bibinfo}[2]{#2}
\providecommand{\BIBentrySTDinterwordspacing}{\spaceskip=0pt\relax}
\providecommand{\BIBentryALTinterwordstretchfactor}{4}
\providecommand{\BIBentryALTinterwordspacing}{\spaceskip=\fontdimen2\font plus
\BIBentryALTinterwordstretchfactor\fontdimen3\font minus \fontdimen4\font\relax}
\providecommand{\BIBforeignlanguage}[2]{{%
\expandafter\ifx\csname l@#1\endcsname\relax
\typeout{** WARNING: IEEEtran.bst: No hyphenation pattern has been}%
\typeout{** loaded for the language `#1'. Using the pattern for}%
\typeout{** the default language instead.}%
\else
\language=\csname l@#1\endcsname
\fi
#2}}
\providecommand{\BIBdecl}{\relax}
\BIBdecl

\bibitem{deng2020self}
X.~Deng, Y.~Xiang, A.~Mousavian, C.~Eppner, T.~Bretl, and D.~Fox, ``Self-supervised 6d object pose estimation for robot manipulation,'' in \emph{2020 IEEE International Conference on Robotics and Automation (ICRA)}.\hskip 1em plus 0.5em minus 0.4em\relax IEEE, 2020, pp. 3665--3671.

\bibitem{cai2024gs}
D.~Cai, J.~Heikkil{\"a}, and E.~Rahtu, ``Gs-pose: Cascaded framework for generalizable segmentation-based 6d object pose estimation,'' \emph{arXiv preprint arXiv:2403.10683}, 2024.

\bibitem{marchand2015pose}
E.~Marchand, H.~Uchiyama, and F.~Spindler, ``Pose estimation for augmented reality: a hands-on survey,'' \emph{IEEE transactions on visualization and computer graphics}, vol.~22, no.~12, pp. 2633--2651, 2015.

\bibitem{su2019deep}
Y.~Su, J.~Rambach, N.~Minaskan, P.~Lesur, A.~Pagani, and D.~Stricker, ``Deep multi-state object pose estimation for augmented reality assembly,'' in \emph{2019 IEEE International Symposium on Mixed and Augmented Reality Adjunct (ISMAR-Adjunct)}.\hskip 1em plus 0.5em minus 0.4em\relax IEEE, 2019, pp. 222--227.

\bibitem{lowe1999object}
D.~G. Lowe, ``Object recognition from local scale-invariant features,'' in \emph{Proceedings of the seventh IEEE international conference on computer vision}, vol.~2.\hskip 1em plus 0.5em minus 0.4em\relax Ieee, 1999, pp. 1150--1157.

\bibitem{lepetit2005monocular}
V.~Lepetit, P.~Fua \emph{et~al.}, ``Monocular model-based 3d tracking of rigid objects: A survey,'' \emph{Foundations and Trends{\textregistered} in Computer Graphics and Vision}, vol.~1, no.~1, pp. 1--89, 2005.

\bibitem{hinterstoisser2013model}
S.~Hinterstoisser, V.~Lepetit, S.~Ilic, S.~Holzer, G.~Bradski, K.~Konolige, and N.~Navab, ``Model based training, detection and pose estimation of texture-less 3d objects in heavily cluttered scenes,'' in \emph{Computer Vision--ACCV 2012: 11th Asian Conference on Computer Vision, Daejeon, Korea, November 5-9, 2012, Revised Selected Papers, Part I 11}.\hskip 1em plus 0.5em minus 0.4em\relax Springer, 2013, pp. 548--562.

\bibitem{huang2023voxposer}
W.~Huang, C.~Wang, R.~Zhang, Y.~Li, J.~Wu, and L.~Fei-Fei, ``Voxposer: Composable 3d value maps for robotic manipulation with language models,'' \emph{arXiv preprint arXiv:2307.05973}, 2023.

\bibitem{driess2023palm}
D.~Driess, F.~Xia, M.~S. Sajjadi, C.~Lynch, A.~Chowdhery, B.~Ichter, A.~Wahid, J.~Tompson, Q.~Vuong, T.~Yu \emph{et~al.}, ``Palm-e: An embodied multimodal language model,'' \emph{arXiv preprint arXiv:2303.03378}, 2023.

\bibitem{long2024discuss}
Y.~Long, X.~Li, W.~Cai, and H.~Dong, ``Discuss before moving: Visual language navigation via multi-expert discussions,'' in \emph{2024 IEEE International Conference on Robotics and Automation (ICRA)}.\hskip 1em plus 0.5em minus 0.4em\relax IEEE, 2024, pp. 17\,380--17\,387.

\bibitem{lin2024sam}
J.~Lin, L.~Liu, D.~Lu, and K.~Jia, ``Sam-6d: Segment anything model meets zero-shot 6d object pose estimation,'' in \emph{Proceedings of the IEEE/CVF Conference on Computer Vision and Pattern Recognition}, 2024, pp. 27\,906--27\,916.

\bibitem{li2018deepim}
Y.~Li, G.~Wang, X.~Ji, Y.~Xiang, and D.~Fox, ``Deepim: Deep iterative matching for 6d pose estimation,'' in \emph{Proceedings of the European Conference on Computer Vision (ECCV)}, 2018, pp. 683--698.

\bibitem{li2019cdpn}
Z.~Li, G.~Wang, and X.~Ji, ``Cdpn: Coordinates-based disentangled pose network for real-time rgb-based 6-dof object pose estimation,'' in \emph{Proceedings of the IEEE/CVF international conference on computer vision}, 2019, pp. 7678--7687.

\bibitem{he2020pvn3d}
Y.~He, W.~Sun, H.~Huang, J.~Liu, H.~Fan, and J.~Sun, ``Pvn3d: A deep point-wise 3d keypoints voting network for 6dof pose estimation,'' in \emph{Proceedings of the IEEE/CVF conference on computer vision and pattern recognition}, 2020, pp. 11\,632--11\,641.

\bibitem{he2021ffb6d}
Y.~He, H.~Huang, H.~Fan, Q.~Chen, and J.~Sun, ``Ffb6d: A full flow bidirectional fusion network for 6d pose estimation,'' in \emph{Proceedings of the IEEE/CVF conference on computer vision and pattern recognition}, 2021, pp. 3003--3013.

\bibitem{su2022zebrapose}
Y.~Su, M.~Saleh, T.~Fetzer, J.~Rambach, N.~Navab, B.~Busam, D.~Stricker, and F.~Tombari, ``Zebrapose: Coarse to fine surface encoding for 6dof object pose estimation,'' in \emph{Proceedings of the IEEE/CVF Conference on Computer Vision and Pattern Recognition}, 2022, pp. 6738--6748.

\bibitem{wang2019densefusion}
C.~Wang, D.~Xu, Y.~Zhu, R.~Mart{\'\i}n-Mart{\'\i}n, C.~Lu, L.~Fei-Fei, and S.~Savarese, ``Densefusion: 6d object pose estimation by iterative dense fusion,'' in \emph{Proceedings of the IEEE/CVF conference on computer vision and pattern recognition}, 2019, pp. 3343--3352.

\bibitem{wang2021gdr}
G.~Wang, F.~Manhardt, F.~Tombari, and X.~Ji, ``Gdr-net: Geometry-guided direct regression network for monocular 6d object pose estimation,'' in \emph{Proceedings of the IEEE/CVF Conference on Computer Vision and Pattern Recognition}, 2021, pp. 16\,611--16\,621.

\bibitem{xiang2017posecnn}
Y.~Xiang, T.~Schmidt, V.~Narayanan, and D.~Fox, ``Posecnn: A convolutional neural network for 6d object pose estimation in cluttered scenes,'' \emph{arXiv preprint arXiv:1711.00199}, 2017.

\bibitem{wang2019normalized}
H.~Wang, S.~Sridhar, J.~Huang, J.~Valentin, S.~Song, and L.~J. Guibas, ``Normalized object coordinate space for category-level 6d object pose and size estimation,'' in \emph{Proceedings of the IEEE/CVF Conference on Computer Vision and Pattern Recognition}, 2019, pp. 2642--2651.

\bibitem{song2016deep}
S.~Song and J.~Xiao, ``Deep sliding shapes for amodal 3d object detection in rgb-d images,'' in \emph{Proceedings of the IEEE conference on computer vision and pattern recognition}, 2016, pp. 808--816.

\bibitem{qi2018frustum}
C.~R. Qi, W.~Liu, C.~Wu, H.~Su, and L.~J. Guibas, ``Frustum pointnets for 3d object detection from rgb-d data,'' in \emph{Proceedings of the IEEE conference on computer vision and pattern recognition}, 2018, pp. 918--927.

\bibitem{chen2016monocular}
X.~Chen, K.~Kundu, Z.~Zhang, H.~Ma, S.~Fidler, and R.~Urtasun, ``Monocular 3d object detection for autonomous driving,'' in \emph{Proceedings of the IEEE conference on computer vision and pattern recognition}, 2016, pp. 2147--2156.

\bibitem{mousavian20173d}
A.~Mousavian, D.~Anguelov, J.~Flynn, and J.~Kosecka, ``3d bounding box estimation using deep learning and geometry,'' in \emph{Proceedings of the IEEE conference on Computer Vision and Pattern Recognition}, 2017, pp. 7074--7082.

\bibitem{xiang2015data}
Y.~Xiang, W.~Choi, Y.~Lin, and S.~Savarese, ``Data-driven 3d voxel patterns for object category recognition,'' in \emph{Proceedings of the IEEE conference on computer vision and pattern recognition}, 2015, pp. 1903--1911.

\bibitem{tian2020shape}
M.~Tian, M.~H. Ang, and G.~H. Lee, ``Shape prior deformation for categorical 6d object pose and size estimation,'' in \emph{Computer Vision--ECCV 2020: 16th European Conference, Glasgow, UK, August 23--28, 2020, Proceedings, Part XXI 16}.\hskip 1em plus 0.5em minus 0.4em\relax Springer, 2020, pp. 530--546.

\bibitem{leiter2024chatgpt}
C.~Leiter, R.~Zhang, Y.~Chen, J.~Belouadi, D.~Larionov, V.~Fresen, and S.~Eger, ``Chatgpt: A meta-analysis after 2.5 months,'' \emph{Machine Learning with Applications}, vol.~16, p. 100541, 2024.

\bibitem{kirillov2023segment}
A.~Kirillov, E.~Mintun, N.~Ravi, H.~Mao, C.~Rolland, L.~Gustafson, T.~Xiao, S.~Whitehead, A.~C. Berg, W.-Y. Lo \emph{et~al.}, ``Segment anything,'' in \emph{Proceedings of the IEEE/CVF International Conference on Computer Vision}, 2023, pp. 4015--4026.

\bibitem{zhao2023survey}
W.~X. Zhao, K.~Zhou, J.~Li, T.~Tang, X.~Wang, Y.~Hou, Y.~Min, B.~Zhang, J.~Zhang, Z.~Dong \emph{et~al.}, ``A survey of large language models,'' \emph{arXiv preprint arXiv:2303.18223}, 2023.

\bibitem{nguyen2024gigapose}
V.~N. Nguyen, T.~Groueix, M.~Salzmann, and V.~Lepetit, ``Gigapose: Fast and robust novel object pose estimation via one correspondence,'' in \emph{Proceedings of the IEEE/CVF Conference on Computer Vision and Pattern Recognition}, 2024, pp. 9903--9913.

\bibitem{labbe2022megapose}
Y.~Labb{\'e}, L.~Manuelli, A.~Mousavian, S.~Tyree, S.~Birchfield, J.~Tremblay, J.~Carpentier, M.~Aubry, D.~Fox, and J.~Sivic, ``Megapose: 6d pose estimation of novel objects via render \& compare,'' \emph{arXiv preprint arXiv:2212.06870}, 2022.

\bibitem{wen2024foundationpose}
B.~Wen, W.~Yang, J.~Kautz, and S.~Birchfield, ``Foundationpose: Unified 6d pose estimation and tracking of novel objects,'' in \emph{Proceedings of the IEEE/CVF Conference on Computer Vision and Pattern Recognition}, 2024, pp. 17\,868--17\,879.

\bibitem{yan2024gs}
C.~Yan, D.~Qu, D.~Xu, B.~Zhao, Z.~Wang, D.~Wang, and X.~Li, ``Gs-slam: Dense visual slam with 3d gaussian splatting,'' in \emph{Proceedings of the IEEE/CVF Conference on Computer Vision and Pattern Recognition}, 2024, pp. 19\,595--19\,604.

\bibitem{engel2014lsd}
J.~Engel, T.~Sch{\"o}ps, and D.~Cremers, ``Lsd-slam: Large-scale direct monocular slam,'' in \emph{European conference on computer vision}.\hskip 1em plus 0.5em minus 0.4em\relax Springer, 2014, pp. 834--849.

\bibitem{mur2017orb}
R.~Mur-Artal and J.~D. Tard{\'o}s, ``Orb-slam2: An open-source slam system for monocular, stereo, and rgb-d cameras,'' \emph{IEEE transactions on robotics}, vol.~33, no.~5, pp. 1255--1262, 2017.

\bibitem{keetha2024splatam}
N.~Keetha, J.~Karhade, K.~M. Jatavallabhula, G.~Yang, S.~Scherer, D.~Ramanan, and J.~Luiten, ``Splatam: Splat track \& map 3d gaussians for dense rgb-d slam,'' in \emph{Proceedings of the IEEE/CVF Conference on Computer Vision and Pattern Recognition}, 2024, pp. 21\,357--21\,366.

\bibitem{li2023nerf}
F.~Li, S.~R. Vutukur, H.~Yu, I.~Shugurov, B.~Busam, S.~Yang, and S.~Ilic, ``Nerf-pose: A first-reconstruct-then-regress approach for weakly-supervised 6d object pose estimation,'' in \emph{Proceedings of the IEEE/CVF International Conference on Computer Vision}, 2023, pp. 2123--2133.

\bibitem{bay2006surf}
H.~Bay, T.~Tuytelaars, and L.~Van~Gool, ``Surf: Speeded up robust features,'' in \emph{Computer Vision--ECCV 2006: 9th European Conference on Computer Vision, Graz, Austria, May 7-13, 2006. Proceedings, Part I 9}.\hskip 1em plus 0.5em minus 0.4em\relax Springer, 2006, pp. 404--417.

\bibitem{collet2010efficient}
A.~Collet and S.~S. Srinivasa, ``Efficient multi-view object recognition and full pose estimation,'' in \emph{2010 IEEE International Conference on Robotics and Automation}.\hskip 1em plus 0.5em minus 0.4em\relax IEEE, 2010, pp. 2050--2055.

\bibitem{hinterstoisser2011multimodal}
S.~Hinterstoisser, S.~Holzer, C.~Cagniart, S.~Ilic, K.~Konolige, N.~Navab, and V.~Lepetit, ``Multimodal templates for real-time detection of texture-less objects in heavily cluttered scenes,'' in \emph{2011 international conference on computer vision}.\hskip 1em plus 0.5em minus 0.4em\relax IEEE, 2011, pp. 858--865.

\bibitem{rad2017bb8}
M.~Rad and V.~Lepetit, ``Bb8: A scalable, accurate, robust to partial occlusion method for predicting the 3d poses of challenging objects without using depth,'' in \emph{Proceedings of the IEEE international conference on computer vision}, 2017, pp. 3828--3836.

\bibitem{hai2023rigidity}
Y.~Hai, R.~Song, J.~Li, M.~Salzmann, and Y.~Hu, ``Rigidity-aware detection for 6d object pose estimation,'' in \emph{Proceedings of the IEEE/CVF Conference on Computer Vision and Pattern Recognition}, 2023, pp. 8927--8936.

\bibitem{zakharov2018keep}
S.~Zakharov, B.~Planche, Z.~Wu, A.~Hutter, H.~Kosch, and S.~Ilic, ``Keep it unreal: Bridging the realism gap for 2.5 d recognition with geometry priors only,'' in \emph{2018 International Conference on 3D Vision (3DV)}.\hskip 1em plus 0.5em minus 0.4em\relax IEEE, 2018, pp. 1--11.

\bibitem{park2019pix2pose}
K.~Park, T.~Patten, and M.~Vincze, ``Pix2pose: Pixel-wise coordinate regression of objects for 6d pose estimation,'' in \emph{Proceedings of the IEEE/CVF international conference on computer vision}, 2019, pp. 7668--7677.

\bibitem{kerbl20233d}
B.~Kerbl, G.~Kopanas, T.~Leimk{\"u}hler, and G.~Drettakis, ``3d gaussian splatting for real-time radiance field rendering.'' \emph{ACM Trans. Graph.}, vol.~42, no.~4, pp. 139--1, 2023.

\bibitem{matsuki2024gaussian}
H.~Matsuki, R.~Murai, P.~H. Kelly, and A.~J. Davison, ``Gaussian splatting slam,'' in \emph{Proceedings of the IEEE/CVF Conference on Computer Vision and Pattern Recognition}, 2024, pp. 18\,039--18\,048.

\bibitem{yugay2023gaussian}
V.~Yugay, Y.~Li, T.~Gevers, and M.~R. Oswald, ``Gaussian-slam: Photo-realistic dense slam with gaussian splatting,'' \emph{arXiv preprint arXiv:2312.10070}, 2023.

\bibitem{chen2024monogaussianavatar}
Y.~Chen, L.~Wang, Q.~Li, H.~Xiao, S.~Zhang, H.~Yao, and Y.~Liu, ``Monogaussianavatar: Monocular gaussian point-based head avatar,'' in \emph{ACM SIGGRAPH 2024 Conference Papers}, 2024, pp. 1--9.

\bibitem{wang2023gaussianhead}
J.~Wang, J.-C. Xie, X.~Li, F.~Xu, C.-M. Pun, and H.~Gao, ``Gaussianhead: Impressive head avatars with learnable gaussian diffusion,'' \emph{arXiv preprint arXiv:2312.01632}, 2023.

\bibitem{chen2024text}
Z.~Chen, F.~Wang, Y.~Wang, and H.~Liu, ``Text-to-3d using gaussian splatting,'' in \emph{Proceedings of the IEEE/CVF Conference on Computer Vision and Pattern Recognition}, 2024, pp. 21\,401--21\,412.

\bibitem{tang2023dreamgaussian}
J.~Tang, J.~Ren, H.~Zhou, Z.~Liu, and G.~Zeng, ``Dreamgaussian: Generative gaussian splatting for efficient 3d content creation,'' \emph{arXiv preprint arXiv:2309.16653}, 2023.

\bibitem{yi2023gaussiandreamer}
T.~Yi, J.~Fang, G.~Wu, L.~Xie, X.~Zhang, W.~Liu, Q.~Tian, and X.~Wang, ``Gaussiandreamer: Fast generation from text to 3d gaussian splatting with point cloud priors,'' \emph{arXiv preprint arXiv:2310.08529}, 2023.

\bibitem{kehl2017ssd}
W.~Kehl, F.~Manhardt, F.~Tombari, S.~Ilic, and N.~Navab, ``Ssd-6d: Making rgb-based 3d detection and 6d pose estimation great again,'' in \emph{Proceedings of the IEEE international conference on computer vision}, 2017, pp. 1521--1529.

\bibitem{karnati2021lienet}
M.~Karnati, A.~Seal, A.~Yazidi, and O.~Krejcar, ``Lienet: A deep convolution neural network framework for detecting deception,'' \emph{IEEE transactions on cognitive and developmental systems}, vol.~14, no.~3, pp. 971--984, 2021.

\bibitem{cai2020reconstruct}
M.~Cai and I.~Reid, ``Reconstruct locally, localize globally: A model free method for object pose estimation,'' in \emph{Proceedings of the IEEE/CVF Conference on Computer Vision and Pattern Recognition}, 2020, pp. 3153--3163.

\bibitem{zakharov2019dpod}
S.~Zakharov, I.~Shugurov, and S.~Ilic, ``Dpod: 6d pose object detector and refiner,'' in \emph{Proceedings of the IEEE/CVF international conference on computer vision}, 2019, pp. 1941--1950.

\bibitem{peng2019pvnet}
S.~Peng, Y.~Liu, Q.~Huang, X.~Zhou, and H.~Bao, ``Pvnet: Pixel-wise voting network for 6dof pose estimation,'' in \emph{Proceedings of the IEEE/CVF conference on computer vision and pattern recognition}, 2019, pp. 4561--4570.

\bibitem{sun2022onepose}
J.~Sun, Z.~Wang, S.~Zhang, X.~He, H.~Zhao, G.~Zhang, and X.~Zhou, ``Onepose: One-shot object pose estimation without cad models,'' in \emph{Proceedings of the IEEE/CVF Conference on Computer Vision and Pattern Recognition}, 2022, pp. 6825--6834.

\bibitem{he2022onepose++}
X.~He, J.~Sun, Y.~Wang, D.~Huang, H.~Bao, and X.~Zhou, ``Onepose++: Keypoint-free one-shot object pose estimation without cad models,'' \emph{Advances in Neural Information Processing Systems}, vol.~35, pp. 35\,103--35\,115, 2022.

\end{thebibliography}

\end{document}